\def\year{2020}\relax
\documentclass[letterpaper]{article} 
\usepackage{aaai20}  
\usepackage{times}  
\usepackage{helvet} 
\usepackage{courier}  
\usepackage[hyphens]{url}  
\usepackage{graphicx} 
\usepackage{graphicx}  
\usepackage{xcolor}
\usepackage{amssymb}
\usepackage{booktabs} 
\usepackage{amsmath}

\title{TrueLearn: A Family of Bayesian Algorithms to Match\\Lifelong Learners to Open Educational Resources}
\author{
\Large \textbf{Sahan Bulathwela, Mar\'ia P\'erez-Ortiz, Emine Yilmaz and John Shawe-Taylor}\\ 
Department of Computer Science, University College London \\
Gower Street, London WC1E 6BT, UK\\ 
\{m.bulathwela, maria.perez, emine.yilmaz, j.shawe-taylor\}@ucl.ac.uk  
}

\begin{document}

\maketitle

\begin{abstract}
The recent advances in computer-assisted learning systems and the availability of open educational resources today promise a pathway to providing cost-efficient high-quality education to large masses of learners. One of the most ambitious use cases of computer-assisted learning is to build a lifelong learning recommendation system. Unlike short-term courses, lifelong learning presents unique challenges, requiring sophisticated recommendation models that account for a wide range of factors such as background knowledge of learners or novelty of the material while effectively maintaining knowledge states of masses of learners for significantly longer periods of time (ideally, a lifetime). This work presents the foundations towards building a dynamic, scalable and transparent recommendation system for education, modelling learner's knowledge from implicit data in the form of engagement with open educational resources. We i) use a text ontology based on Wikipedia to automatically extract knowledge components of educational resources and, ii) propose a set of online Bayesian strategies inspired by the well-known areas of item response theory and knowledge tracing. Our proposal, TrueLearn, focuses on recommendations for which the learner has enough background knowledge (so they are able to understand and learn from the material), and the material has enough novelty that would help the learner improve their knowledge about the subject and keep them engaged. We further construct a large open educational video lectures dataset and test the performance of the proposed algorithms, which show clear promise towards building an effective educational recommendation system.

\end{abstract}

\section{Introduction}
One-on-one tutoring has shown learning gains 
of the order of two standard deviations \cite{Corbett2001}. 
Machine learning now promises to provide such benefits of 
high quality personalised teaching to anyone in the world in a cost effective manner~\cite{Piech2015}.
Meanwhile, Open Educational Resources (OERs), defined as teaching, learning and research material  available  in  the  public  domain  or published under an open license \cite{unesco1}, are growing at a very fast pace.
This work is scoped at creating a personalisation model to identify and recommend the most suitable educational materials, assisting learners on their personal learning pathway to achieve impactful learning outcomes.

Personalised learning systems usually consist of two components \cite{Lan2014}: (i) {learning analytics},
that capture the dynamic
learner's knowledge state and (ii) {content analytics}, that extract  characteristics of the learning resource, such as knowledge components covered and resource quality/difficulty. In the context of learning analytics, the assessment and learning science communities 
aim to assess learner's knowledge at a specific time point (e.g. during a test). 
Historically, content analytics
have been provided by human experts. Although expert labelling appears to be sensible, the rapid growth of educational resources demands for scalable and automatic annotation.

While excelling on the personalisation front, there are other features that the ideal educational recommendation system should have. We design our system with these features in mind: 
(i) {Cross-Modality} and (ii) {Cross-linguality} are vital to identifying and recommending educational resources across different modalities and languages. (iii) {Transparency} empowers the learners by building trust while supporting the learner's metacognition processes such as planning, monitoring and reflection (e.g. Open Learner Models \cite{Bull2016}). (iv) {Scalability} ensures that a high quality learning experience can be provided to large masses of learners over long periods of time. (v) {Data efficiency} enables the system to use all available data efficiently (e.g. learning from implicit engagement rather than sourcing explicit feedback).

This paper proposes {a family of Bayesian strategies aimed at providing educational recommendations to learners} using learner's implicit engagement with OERs. To do so, we use an ontology based on Wikipedia to extract content analytics from the text representation of educational resources. Our objective is to develop an adaptive and scalable system that can recommend suitably complex material and is transparent to learners. 
Our approach differs from previous work in several ways: (i) It can be applied in situations where explicit feedback about the learner's knowledge is unavailable, as tends to be the case in informal lifelong learning; and (ii) we focus on recovering learner's knowledge, as opposed to previous work on recommender systems which mostly focuses on leveraging user interests. 
We test the different models and assumptions using a VideoLectures.net dataset composed of 18,933 learners and 248,643 view log
entries, with promising results.

\section{Related Work} \label{topic:related_work}


Modelling learners and resources is of fundamental importance to all adaptive educational systems.
Most literature 
in the adaptive educational systems domain
focuses on estimating learner's knowledge based on test answers,
modelling the learner at a static point in time and assessing a limited set of skills (in most cases, individual skills). However, for lifelong learning, a wider range of skills has to be modelled over long spans of time and prior research in this area is surprisingly scarce. We review in this section content and learning analytics, focusing for the latter on {Item Response Theory} (IRT) and {Knowledge Tracing} (KT), from which we draw inspiration.


\subsection{Content Analytics: Knowledge Components}

Content representations play a key role in recommending relevant  materials to learners. In education, this entails extracting atomic units of learnable concepts that are contained in a learning resource. We refer to these concepts as \textbf{Knowledge Components (KCs)} that can be learned and mastered. However, KC extraction is challenging and expert labelling is the most commonly used approach. Although automated techniques have been proposed \cite{Lindsey2014}, these usually rely on partial expert labelling or the use of unsupervised learning approaches, which are complex to tune. Recent advances in deep learning have also led to the proposal of deep models that learn latent KCs \cite{Piech2015}. However, these deep representations make the interpretability of the learner's models and the resource representations more challenging. \emph{Wikification}, an entity linking recent approach, looks promising towards automatically extracting explainable KCs. Wikification identifies Wikipedia concepts present in the resource by connecting natural text to Wikipedia articles via entity linking \cite{wikifier}. This approach avoids expensive expert labelling while providing an ontology of domain-agnostic humanly interpretable KCs. However, Wikipedia KCs may not be as accurate as those carefully crafted by education experts. 





\subsection{Item Response Theory (IRT) and Matchmaking} \label{topic:kt}

IRT \cite{Rasch1960} focuses on designing, analysing and scoring ability tests by modelling learner's knowledge and question difficulty. However, IRT does not consider changes in knowledge over time. The simplest model, known as Rasch model \cite{Rasch1960}, proposes to compute the probability of scoring a correct answer as a function of the learner's skill $\theta_\ell$ and the difficulty of the question/resource $d_r$: 
\begin{equation}
    P(\texttt{correct answer} | \theta_\ell, d_r) = f(\theta_\ell - d_r),
\end{equation} where $f$ is usually a logistic function. This idea has been extended to algorithms such as Elo \cite{elo}, to rank chess players based on their game outcomes, where instead of having learners and resources, two players compete. The well-known TrueSkill algorithm \cite{trueskill} improves this skill learning setting using a Bayesian approach, allowing teams of players to compete and adding a dynamic component to update player skills over time. Previous work has proposed the use of Elo-based algorithms for modelling learners \cite{Pelanek2017}, because of its similarity to the Rasch model and its computationally light online version. More recent works such as Knowledge Tracing Machines, that have a direct relationship with IRT have shown that it can outperform state-of-the-art deep learning models such as Deep Knowledge Tracing in some cases \cite{vie2019knowledge}.

\subsection{Knowledge Tracing (KT)} \label{topic:kt}

KT \cite{corbett1994knowledge} is one of the most widespread models used in intelligent tutoring systems, the main difference with IRT being that question difficulty is not considered.  It aims to estimate knowledge acquisition as a function of practice opportunities. Numerous variants of KT are emerging, e.g. enabling individualisation \cite{yudelson2013individualized}. More recently, Deep Knowledge Tracing \cite{Piech2015} has shown improvement over KT. However, the challenges in interpretability of the learned KCs can be seen as a major drawback of deep KT.

\subsection{Educational Recommender Systems} \label{edurec}

Although conventional recommendation approaches (e.g. collaborative filtering) have been proposed for education \cite{bobadilla2009collaborative}, educational recommenders have associated unique challenges, 
These challenges stem from the objective of bringing learners closer to their goals in the most effective way and specifically comprise: i) Identifying learners interests and learning goals, as these can significantly affect their motivation \cite{Salehi2014}; ii) identifying the dynamic background knowledge of learners, the topics covered in a resource and the prerequisites necessary for benefiting from a learning material; iii) recommending novel and impactful materials to learners and planning learning trajectories suitable for learners; and iv) accounting for how different resource quality factors impact how engaging an educational resource may be to the general population \cite{Guo_vid_prod}.

In the recent years, hybrid approaches \cite{hybrid_rec,Salehi2014} and deep learning methods \cite{goal_based_edrec} have been proposed to improve educational recommendation systems, incorporating additional handcrafted information such as learning trajectories and goals \cite{bauman2018recommending}. However, much still remains to be done. Most of these approaches rely on manually handcrafting learning trajectories, which is highly domain specific and hard to scale.
Moreover, hybrid approaches do not address novelty and deep learning methods suffer from a lack of transparency of the learned representations. These challenges motivate the development of accurate, scalable, simple, and transparent educational recommendation systems, which is our aim with this work. 

\section{Modelling Implicit Educational Engagement}

Implicit feedback has been used for building recommender systems for nearly two decades with great success \cite{Oard1998ImplicitFF,Jannach2018RecommendingBO}, as an alternative to explicit ratings, which have a high cognitive load on users and are generally sparse. 
For videos, normalised {watch time} is commonly used as an implicit proxy for engagement \cite{Covington2016,Guo_vid_prod},
shown to increase the likelihood of achieving better learning outcomes \cite{pardos2014affective,carini2006student,ramesh2014learning,lan2017behavior}. 
We identify several factors \cite{Bulathwela2020} in the learning science literature that influence engagement: i) background knowledge \cite{yudelson2013individualized}, ii) novelty of the material \cite{Drachsler:edurec}, iii) learners interests or learning goals \cite{Salehi2014} and iv) marginal engagement or quality \cite{Lane10,quality_features} of learning resources. This paper focuses on i) and ii), as a first step towards building an integrative recommendation system that accounts for all four.  


We first present two relatively na\"ive baselines for modeling engagement. 
Secondly, we describe how we adapt two approaches from the literature, namely TrueSkill and KT. Then, we propose an extension of TrueSkill, defining educational engagement as a function of background knowledge and novelty. 
This is, the proposed system recommends resources for which the learner has the necessary background knowledge but there is novelty. 
To be able to handle large-scale scenarios, our work focuses on online learning solutions that are massively parallelisable, prioritising models that can be run per learner, for simplicity and transparency.


\begin{figure*}[ht!]
\centering
\includegraphics[width=\textwidth]{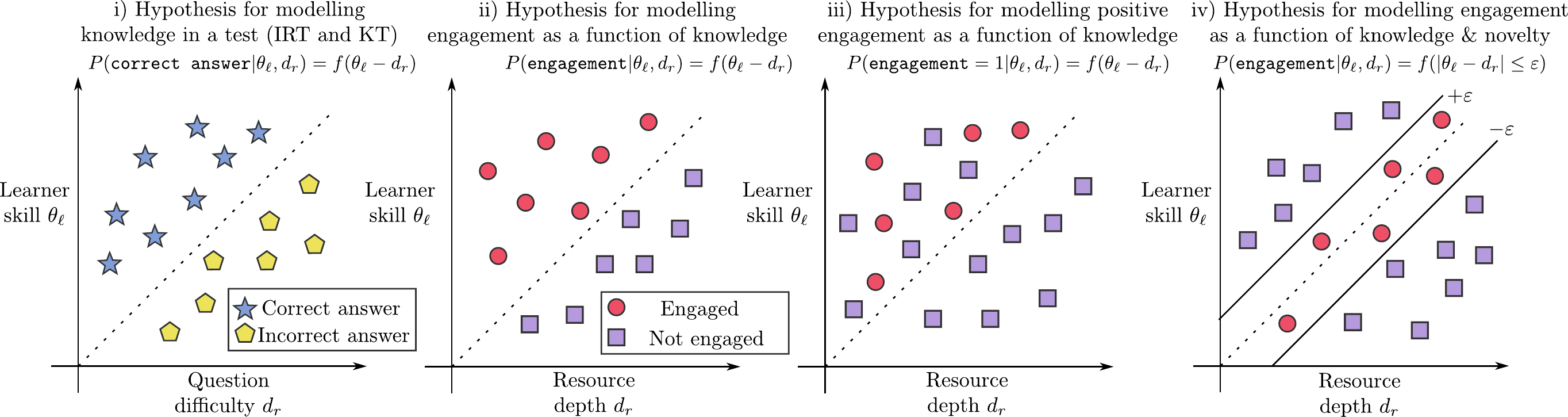}
\caption{Graphical representation of different assumptions that can be made when modelling learner's knowledge. The methods tested in this paper are set to test these four hypotheses. }
\label{fig:hypotheses}
\end{figure*}

\subsection{Problem Formulation and Assumptions}
Consider a learning environment in which a learner $\ell$ interacts with a set of educational resources $S_\ell \subset \{r_1, \ldots, r_Q\}$ over a period of $T= (1,\ldots,t)$ time steps, $Q$ being the total of resources in the system.
A resource $r_i$ is characterised by a set of top KCs or topics $K_{r_i} \subset \{1, \ldots, N \}$ ($N$ being the total of KCs considered by the system) and the depth of coverage $d_{r_i}$  of those. The key idea is to model the probability of engagement $e_{\ell, r_i}^{t} \in \{ 1, -1\}$ between learner $\ell$ and resource $r_i$ at time $t$ as a function of the learner skill $\theta^t_\ell$ and 
resource representation $d_{r_i}$ for the top KCs covered $K_{r_i}$.
According to Bayes rule the posterior distribution is proportional to:
\begin{equation}
 P({\theta}^t_\ell | e^{t}_{\ell,r_i}, K_{r_i}, d_{r_i} ) \propto P( e^{t}_{\ell,r_i} | \theta^t_\ell, K_{r_i}, d_{r_i}) \cdot P(\theta^t_\ell).
\end{equation}

\figurename{ \ref{fig:hypotheses}} shows the intuition behind different hypothesis for modelling one single learner skill. Hypothesis i) shows the assumption made in IRT and KT (both focused on test scoring). This is, if the learner answers correctly to a test, the skill must exceed the difficulty of the question. The boundary of $\theta_\ell - d_r$ is shown using a dotted line in all cases. 
Hypothesis ii) shows the analogue for engagement (as a function of knowledge), i.e. if the learner is engaged, they have enough background to make use of the resource and vice versa. However, we hypothesize that this is very restrictive and that no assumption can be made from the non-engaged cases (the learner might not be engaged for a myriad of reasons, e.g. the learner being advanced in a topic and finding the resource too easy, in which case we can not say that they lack the necessary background). This idea that we can only learn background knowledge from positive engagement is shown in hypothesis iii) and is a common assumption when learning from implicit feedback \cite{Jannach2018RecommendingBO}. The last plot (hypothesis iv)) shows the combination of knowledge and novelty: if the learner is engaged, they must have the appropriate background and the content must also be novel to them (i.e. neither too easy nor too difficult). We introduce $\varepsilon$ as the engagement margin. 

\subsection{Na\"ive Baselines for Engagement}\label{sec:baselines}

Since learning from 
engagement with educational resources is a novel research area we could not find suitable baselines to compare against. 
Our first contribution is to propose two relatively na\"ive baselines: i) persistence, which assumes a static behaviour for all users $P(e_{\ell, \cdot}^t) = P(e_{\ell, \cdot}^{t-1})$, where $(\cdot)$ indicates any resource, i.e. if the learner is engaged, they will remain engaged and vice versa; ii) majority of user engagement, which predicts future engagement based solely on mean past engagement of users, i.e. $P(e_{\ell, \cdot}^t) = \frac{1}{n} \cdot \sum_{i=1}^{t-1} P(e_{\ell, \cdot}^{i})$.
The persistence baseline assumes a common model for both learners and resources. The majority baseline assumes differences between users, and disregards resource differences. 

\subsection{Modelling Knowledge}\label{sec:knowledge}
Our second contribution is to
extend the most well-known approaches for modelling skills/knowledge: TrueSkill \cite{trueskill} and KT \cite{corbett1994knowledge}. Our proposed model, {TrueLearn}, is inspired by TrueSkill with regard to representing {and learning skills}. We use TrueSkill as the foundation for TrueLearn because we observe that it could predict engagement better than KT in preliminary experiments. 

In TrueSkill, each player $\ell$  is assumed to have an unknown real skill $\theta^t_\ell \in \mathbb{R}$, exhibiting a performance $p^t_\ell$ drawn according to $p(p^t_\ell| \theta_\ell^t) = \mathcal{N}(p^t_\ell; \theta_\ell^t, \beta^2)$ with fixed variance $\beta^2$. The outcome of the game $y^t_{jz}$ between two players $\ell_j$ and $\ell_z$ (in our case learner $\ell$ and resource $r_i$) is modelled as:
\begin{equation}
P(p^t_{\ell_j} > p^t_{\ell_z} | \theta^t_{\ell_j},\theta^t_{\ell_z}) := \Phi  \left ( \frac{ \theta^t_{\ell_j} - \theta^t_{\ell_z} }{\sqrt{2}\beta} \right ),
\end{equation}
where $\Phi$ is the cumulative density of a zero-mean unit variance Gaussian. 
To adapt TrueSkill, we consider 
four approaches. The first two, referred to as \textbf{Vanilla TrueSkill} and \textbf{Vanilla TrueSkill Video}, represent the original TrueSkill algorithm and model a single skill for the learner $\theta_\ell$ and the depth of the resource $d_{r_i}$ as two players competing.
The engagement label is used as the outcome of the game, meaning that if the learner is engaged, the skill of the learner is equal or larger than the depth of the resource $P(e^t_{\ell, r_i}) = P(p^t_{\ell} > p_{r_i})$.
The main difference between Vanilla TrueSkill and Vanilla TrueSkill Video is that the former considers the basic atomic unit of knowledge to be a fragment of a video (as the rest of approaches in this work), whereas the video version learns only one skill for the whole resource. This partitioning of lectures into multiple fragments was done to capture knowledge acquisition in a finer grain level, which means that for all methods tested in this paper (with exception of Vanilla TrueSkill Video) a fragment of the lecture is considered as a resource $d_{r_i}$. The comparison with the video version, although not completely fair, was intended to test the assumption that the topics within a lecture are likely to be correlated and can be treated as one.

As opposed to the Vanilla TrueSkill versions, we wish to model multiple skills for a learner: $\boldsymbol{\theta}_\ell = (\theta_{\ell,1}, \ldots, \theta_{\ell,N})$.
TrueSkill also allows for teams, assuming that the performance of a team is the sum of the individual performance of its players. 
The 
third
version of TrueSkill that we consider (our first 
novel
proposal, \textbf{TrueLearn dynamic-depth}) is based on two teams playing, where both the learner and resource are represented as a "team of skills":
\begin{equation}
    p^t_\ell = \sum_{h \in K_{r_i}} \theta^t_{\ell,h}, \; \;\; \; p_{r_i} = \sum_{h \in K_{r_i}} d_{r_i,h}.
\end{equation} 
Knowledge components $K_z$ thus define teams. We consider this approach rather than assuming that each individual skill has to win over its associated KC depth because we observed that most KCs represent related topics. A similar approach using Elo system and knowledge for only one skill was considered in \cite{Pelanek2017}. For the 
fourth 
model (named \textbf{TrueLearn fixed-depth}), we use a similar approach but fix one branch to the observed knowledge depth, using  text cosine similarity defined in the next section. 

Unlike TrueSkill, KT uses Bernoulli variables to model skills $\theta^t_{\ell,h} \sim \texttt{Bernoulli}(\pi^t_{\ell,h})$, assuming that a learner $\ell$ would have either mastered a skill or not (represented by probability $\pi^t_{\ell,h}$). Since the objective of KT is not to model learning but to capture the state of mastery at given time, 
KT considers that once a learner has mastered a skill it cannot be unlearnt.
For the extension of KT (named \textbf{Multi skill KT}), we also consider multiple skills. 
Skills are initialised using a $\texttt{Bernoulli}(0.5)$ prior, assuming that the latent skill is equally likely to be mastered than not. A noise factor is also included (similarly to $\beta$ in TrueSkill). This reformulation is inspired by \cite{bishopsnewbook}.

\begin{figure*}[ht!]
\centering
\includegraphics[width=0.95\textwidth]{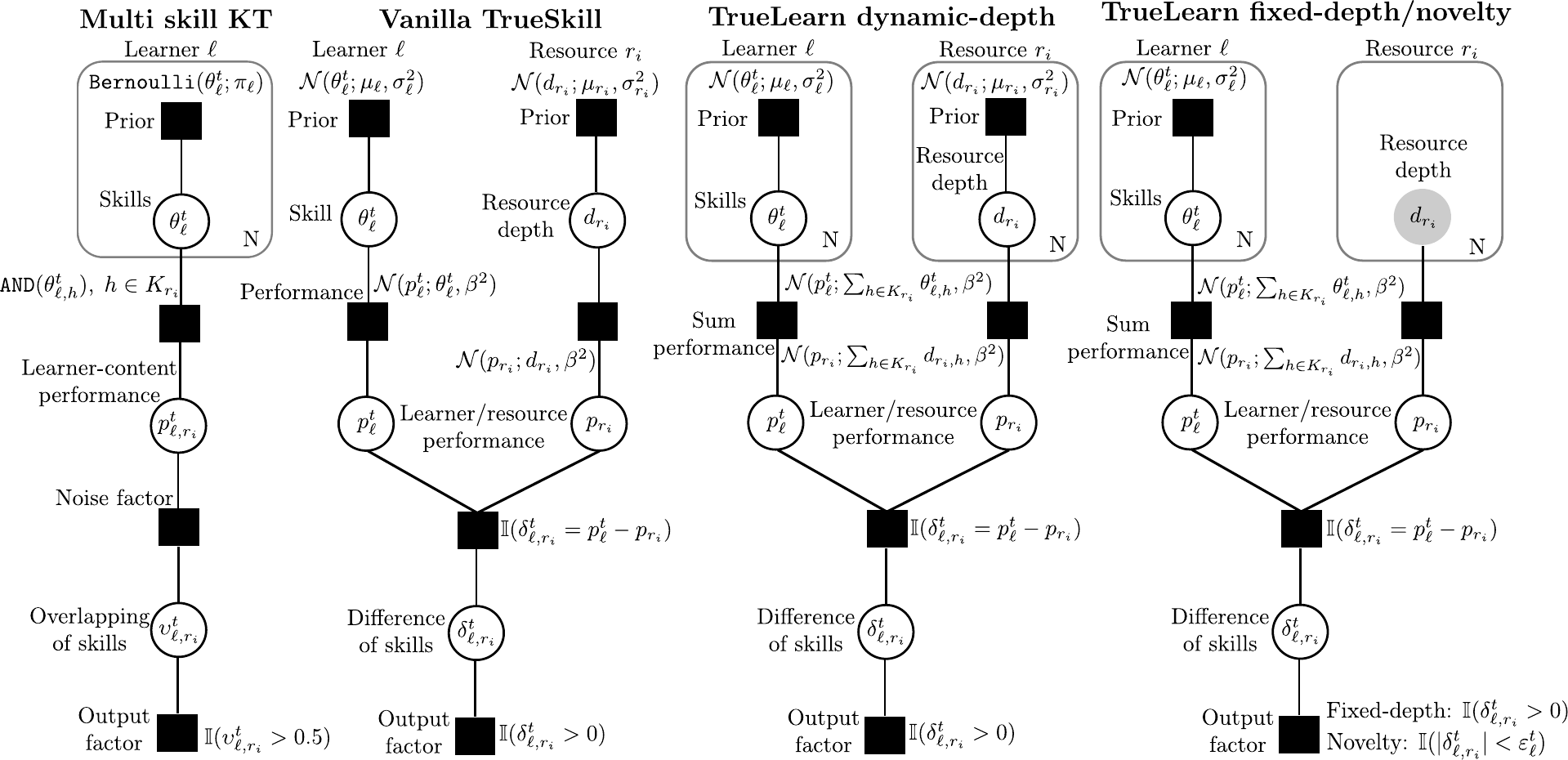}
\caption{Factor graph for Multi skill KT, Vanilla TrueSkill and different versions of the TrueLearn model. Plates represent groups of variables ($N$ Wikipedia topics).}
\label{fig:factorgraph1}
\end{figure*}

\figurename{ \ref{fig:factorgraph1}} shows a representation of the factor graphs used for these three models, together with TrueLearn, covered in the next section. A factor graph is a bi-partite graph consisting of variable and factor nodes, shown respectively with circles and squares. Gray filled circles represent observed variables. Message passing is used for inference, where messages are approximated as well as possible through moment matching.
 Since our aim is to report skill estimates in real-time after learner's activity, we use an online
learning scheme referred to as density filtering for all models, where the posterior distribution is used as the prior distribution for the next time instant. 
The models presented here are used to test hypotheses ii) and iii) in \figurename{ \ref{fig:hypotheses}}. 

\subsection{TrueLearn: Extending Knowledge with Novelty}\label{sec:truelearn}

\textbf{TrueLearn Novelty} additionally introduces {novelty}, defined as the degree to which a resource contains new information for the learner. 
Engagement outcomes $e^t_{\ell,r_i}$ between learner $\ell$ and resource $r_i$ are determined in this case as:
\begin{equation}
e^t_{\ell,r_i} := 
\begin{Bmatrix}
+ 1 \; \; \texttt{if} | p^t_{\ell} - p_{r_i} | \leq \varepsilon^t_\ell\\
-1 \; \; \texttt{otherwise},
    \end{Bmatrix}
\end{equation}
where the parameter $\varepsilon^t_\ell>0$ is referred to as the engagement margin and is learner dependent. This represents the idea that both the learner and resource must be found in a similar knowledge state for the learner to be engaged (hypothesis iv) in \figurename{ \ref{fig:hypotheses}}).
This engagement margin $\varepsilon^t_\ell$ is set counting the fraction of engaged outcomes for a learner and relating the margin to the probability of engagement by: 
\begin{equation}
    P(e^t_{\ell,\cdot}) = \Phi \left ( \frac{\varepsilon^t_\ell}{\sqrt{|K_{r_i}|}\beta} \right ) - \Phi \left ( \frac{-\varepsilon^t_\ell}{\sqrt{|K_{r_i}|}\beta} \right ).
\end{equation}
The model is shown in \figurename{ \ref{fig:factorgraph1}}. 
Here, we learn learner's knowledge from positive and negative engagement.
The function represented by the factor graph
is the joint distribution $p(\theta^t_\ell, p^t_\ell, p_{r_i} | e^t_{\ell,r_i}, K_{r_i}, d_{r_i})$, given by the product of all the
functions associated with each factor. The posterior $p(\theta^t_\ell | e^t_{\ell,r_i}, K_{r_i}, d_{r_i})$ is computed from the joint distribution integrating the performances $p^t_\ell$ and $p_{r_i}$. 
For recommendations, a resource $r_i$ can be ranked using $P(e^t_{\ell,r_i})$.

\paragraph{Dynamics}


So far all models assume a stationary data
distribution and hence in the limit of infinite observations, learning would come to a halt. Like in TrueSkill, we consider a Gaussian drift over skills between time steps given by $p(\theta^t_\ell | \theta^{t-1}_\ell) = \mathcal{N}(\theta_\ell^t; \theta_\ell^{t-1}, \tau^2)$.   This is introduced as an additive variance component in the subsequent prior and is crucial for lifelong learning.  For Multi skill KT,
we increase the uncertainty by moving $\pi_\ell$ in the direction of 0.5 probability in steps of $\tau$. However, our results show no change when adding this dynamic factor. We hypothesize that this is because most user sessions in our dataset are relatively short (e.g. there are only 20 users with more than 200 events and these events contain a wide range of topics covered).

\section{Processing OERs: Wikifier and Dataset}\label{dataset}

We set high importance to leveraging cross-modal and cross-lingual capabilities, as these are vital to processing all types of open educational resources in the real-world. We choose text as 
a generic form of raw representation for resources as the majority of modalities (videos, audio, books, web pages, etc.) can be easily converted to text. From text we extract KCs, together with the coverage depth.

\subsection{Knowledge Representation}

Extracting the knowledge components present in an educational resource is a non-trivial task. 
We use an ontology based on Wikipedia to represent KCs, where each Wikipedia page is considered as an independent and atomic unit of knowledge (i.e. a KC). 
Specifically, we use {Wikifier}\footnote{\url{www.wikifier.org}}, an entity linking technique that annotates resources with relevant Wikipedia concepts \cite{wikifier}. Wikifier identifies Wikipedia concepts that are associated with a document and ranks them. A graph that represents semantic relatedness between the Wikipedia concepts is generated from the document and it is used for ranking concepts. 
It shows improved results upon several state-of-the-art entity linking and disambiguation approaches in a battery of datasets.
We rather treat Wikifier as a state-of-the-art entity linking tool and consider improving or testing it out of the scope of this work.
The algorithm uses two main statistics that are computed for each Wikipedia topic associated with the resource: (i) the \emph{PageRank score}, that represents the authority of a topic within 
the resource 
\cite{pagerank}) and (ii) \emph{the cosine similarity} between the Wikipedia page of the topic and the resource. We use cosine similarity as a proxy for the depth of knowledge covered in the resource. A combination of PageRank and cosine similarity is used to rank the top most relevant Wikipedia topics associated to a resource. 
Each Wikipedia topic is defined as a learnable KC. 
We also divide resources into what we call learnable units (fragments). A resource is then composed of different fragments. We believe this is meaningful for two reasons: (i) it enables recommending fine-grained 
resource fragments suited for the learner's learning path, rather than only whole resources, and (ii) because in many cases the learner might not consume the resource entirely (e.g. a book), and we may want to learn exactly from the different fragments consumed. 

\subsection{Dataset} \label{sec:datasets}

We use data from a popular OER repository to evaluate the performance of the models. The data source consists of users watching video lectures from VideoLectures.Net\footnote{\url{www.videolectures.net}}. The lectures are also accompanied with transcriptions and multiple translations provided by the TransLectures project\footnote{\url{www.translectures.eu}}. 
We use the English transcription of the lecture (or the English translation where the resource is non-English) to annotate the lecture with relevant KCs using Wikifier. We divide the lecture text into multiple fragments of approximately 5,000 characters (equivalent roughly to 5 minutes of lecture).
The choice of video partitioning is motivated by several reasons. The first one is a technical limitation on the number of characters supported by Wikifier. However, we also believe that these partitions allow us to use finer-grain engagement user signals, where our algorithm learns from the specific partitions that the user watched (and the topics covered in those). At the moment we use fixed-length partitioning, but in future work, we aim to use more sensible video partitioning, that comes from the actual topics covered. Using improved partitioning based on topics (as opposed to fixed-length) is likely to further improve the performance of our model. 

Once the fragments are wikified, we rank the topics using a linear combination of pagerank and cosine similarity (further details in the next section) and use the top $k$ ranked topics along with the associated cosine similarity as our feature set.
We define binary engagement $e^t_{\ell,r_i}$ between a learner $\ell$ and a resource $r_i$ as $1$ if the learner watched at least 75\% of the fragment of 5000 characters, and -1 otherwise. This is because we hypothesise that the learner must have consumed approximately the whole fragment to learn significantly from it. Note that user view logs are of learners actively accessing videos, i.e. when engagement is negative the learner has accessed the material but left without spending a significant amount of time on it.

The source dataset consisted of 25,697 lectures as of February 2018 that were categorised into 21 subjects, e.g.~ Data Science, Computer Science, Arts, Physics, etc.
 However, as VideoLectures.net has a heavy presence of Computer Science and Data Science lectures, we restricted the dataset to lectures categorised under Computer Science or Data Science categories only. To create the dataset, we extracted the transcripts of the videos and their viewers' view logs. A total of 402,350 view log entries were found between December 8, 2016 and February 17, 2018. 
 These video lectures are long videos that run for 36 minutes on average and hence discuss a large number of KCs in a single lecture. 

We create three distinct datasets, based on the number of learners and top $k$ topics selected. 
The first two datasets (\texttt{20 learners-10 topics} and \texttt{20 learners-5 topics}) are created using the 20 most active users and 10 and 5 top topics respectively. These $20$ users are associated with 6,613 unique view log entries from 400 different lectures. 
The third dataset (\texttt{All learners-5 topics}) consists of events from all users and is composed of {248,643 view log entries distributed among 18,933 users interacting with 3,884 different lectures}. The 5 highest ranked KCs are used for this dataset. The dataset with 10 topics has 10,524 unique KCs while the other two datasets (20 users and all users) with top 5 ranked topics have 7,948 unique KCs. 

\section{Experiments} \label{topic:experiments}

In this section we present the results obtained for the different datasets presented in the previous section. The experiments are set to validate: i) the use of KT against IRT inspired models (and thus the use of Gaussian and Bernoulli variables), ii) the different components that we propose to predict engagement (knowledge and novelty) and iii) the number of top $k$ topics used to characterise a fragment.

\paragraph{Validating Wikifier:} 
The authors of Wikifier proposed a linear combination of Pagerank and cosine similarity to rank relevant topics for a document. In our work, we identified the best linear combination using a grid search on training data with the Multi skill KT model and F1 score. The linear combination is used strictly for ranking the most relevant Wikipedia topics. Once the top topics are identified, only the cosine similarity is used as a proxy for knowledge depth. Cosine similarity, being the inner product of bag-of-words (TF-IDF) representations of the document and the Wikipedia page of topic, is an intuitive proxy for depth of topic related knowledge covered in a resource.

Analysing the results Wikifier \cite{wikifier} produced for several lectures we hypothesized that neither pagerank nor cosine similarity alone could be used to reliably rank the KCs associated to a resource. Pagerank seemed to be fine-grained and prone to transcript errors. Cosine similarity, on the other hand, resulted in very general topics, such as 'Science', 'Data' or 'Time'. We firstly experimented with a linear combination of these two and manually validated the superior accuracy obtained. Such a linear combination was also proposed by the authors in \cite{wikifier}, however they did not report improvements. We then proceed to test different weights for the linear combination using Multi skill KT and F1. In order to find the linear weights, we executed a grid search where values between [0, 1] were assigned to the weights before training. We concluded that the best results were obtained by weighting pagerank by 0.4 and cosine by 0.6. 
\begin{table*}
  \caption{Weighted average test performance for accuracy, precision, recall and F1. 
  Models labelled with ($\bigtriangleup$) are trained with positive and negative engagement. Models labelled with ($*$) learn multiple skill parameters, one per Wikipedia page.}
  \label{tab:truelearn_performance}
  \centering
  \begin{tabular}{l | rrrr | rrrr | rrrr}
    \toprule
     & \multicolumn{4}{c|}{20 learners-5 topics} & \multicolumn{4}{c|}{20 learners-10 topics}  & \multicolumn{4}{c}{All learners-5 topics} \\
     Algorithm    & Acc. & Prec. & Rec. & F1  & Acc. & Prec. & Rec. & F1   & Acc. & Prec. & Rec. & F1\\
    \midrule
    Na\"ive persistence ($\bigtriangleup$) & 
    \textit{.808} & \textbf{.818} & .819 & .819 & 
    .808 & \textbf{.818} & .819 & \textbf{.819} & 
    \textit{.766} & \textbf{.629} & .625 & .625 \\
    
    Na\"ive majority ($\bigtriangleup$) & 
    \textbf{.821} & \textit{.775} & .842 & .794 &
    .821 & \textit{.775} & .842 & .794 & 
    \textbf{.771} & .559 & .633 & .583 \\
    
    Vanilla TrueSkill ($\bigtriangleup$) &
    .733 & .569 & .591 & .574 &
    .733 & .569 & .591 & .574 &
    .610 & .541 & .472 & .480 \\
    
    Vanilla TrueSkill Video ($\bigtriangleup$) &
    .739 & .734 & \textit{.982} & \textit{.814} &
    .794 & .720 & \textit{.955} & .792 &
    .641 & .608 & \textbf{.837} & \textbf{.679} \\
    
    Multi skill KT ($\bigtriangleup$,$*$) &
    .722 & .748 & .753 & .738 & 
    .714 & .700 & .580 & .631 & 
    .498 & .492 & .188 & .254 \\
    
    Multi skill KT ($*$) &
    .715 & .746 & .776 & .745 &
    .715 & .701 & .588 & .638 & 
    .497 & .491 & .192 & .256 \\
    
     TrueLearn dynamic-depth ($\bigtriangleup$,$*$) &
     .753 & .732 & .880 & .789 & 
     .790 & .703 & .753 & .726 &
     .454 & .530 & .431 & .418 \\
    
     TrueLearn fixed-depth ($\bigtriangleup$,$*$) & 
     \textit{.808} & .770 & .815 & .790 & 
     \textbf{.841} & .733 & .794 & .761 &
     .736 & \textit{.610} & .558 & .573 \\
     
     TrueLearn fixed-depth ($*$) &
     .735 & .734 & .\textbf{984} & .813 &
     .759 & .710 & \textbf{.987} & .783 & 
     .719 & .608 & .686 & .626 \\
     
    TrueLearn Novelty ($\bigtriangleup$,$*$) &
    .793 & .754 & .923 & \textbf{.821} &
    \textit{.828} & .722 & .875 & .784 & 
    .649 & .603 & \textit{.835} & \textit{.677} \\
   
    \bottomrule
  \end{tabular}
\end{table*}

\paragraph{Experimental design and evaluation metrics} Given that we aim to build an online system, we test the different models using a sequential experimental design, where engagement of fragment $t$ is predicted using fragments $1$ to $t-1$. We also use a one hold-out validation approach for hyperparameter tuning where hyperparameters are learned on 70\% of the learners and the model is evaluated on the remaining 30\% with the best hyperparameter combination. Note that we both learn and predict the engagement per fragment.
Since engagement is binary, predictions for each fragment can be assembled into a confusion matrix, from which we compute well-known binary classification metrics such as accuracy, precision, recall and F1-measure. We average these metrics per learner and weight each learner according to their amount of activity in the system. Note that most learners present an imbalanced setting, where they are mostly engaged or disengaged. Because of this, we do not use Accuracy as the main metric, but rather focus on Precision, Recall and F1. 
For all models each user is run separately, except for the original TrueSkill, in which we also need to model the difficulty of content and thus we require all users. 
Regarding initial configurations and hyperparameters, we initialised the initial mean skill of learners to 0 for all reformulations of TrueSkill. We use grid search to find the suitable hyperparameters for the initial variance while keeping $\beta$ constant at 0.5. The search range for the initial variance was [0.1, 2]. For these models, initial hyper parameters are set in the following manner. 
For the original TrueSkill setting (Vanilla TrueSkill), we set the same hyperparameters used in \cite{trueskill}. 
For the reformulations of KT, we run a hyperparameter grid search for the probability values of the noise factor in the range [0, 0.3].
We also tested different combinations of $\tau$ ($0.1, 0.05, 0.01$), the hyperparameter controlling the dynamic factor. However, the results did not changed for different settings. This suggests that the dataset might still be relatively small and sparse for this factor to have an impact.  
The algorithms were developed in python, using MapReduce to parallelise the computation per learner. The code for TrueLearn and all the baselines is available online\footnote{\url{https://github.com/sahanbull/TrueLearn}}.

\subsection{Results}
\begin{figure}[!tbp]
 \centering
 \includegraphics[width=\columnwidth]{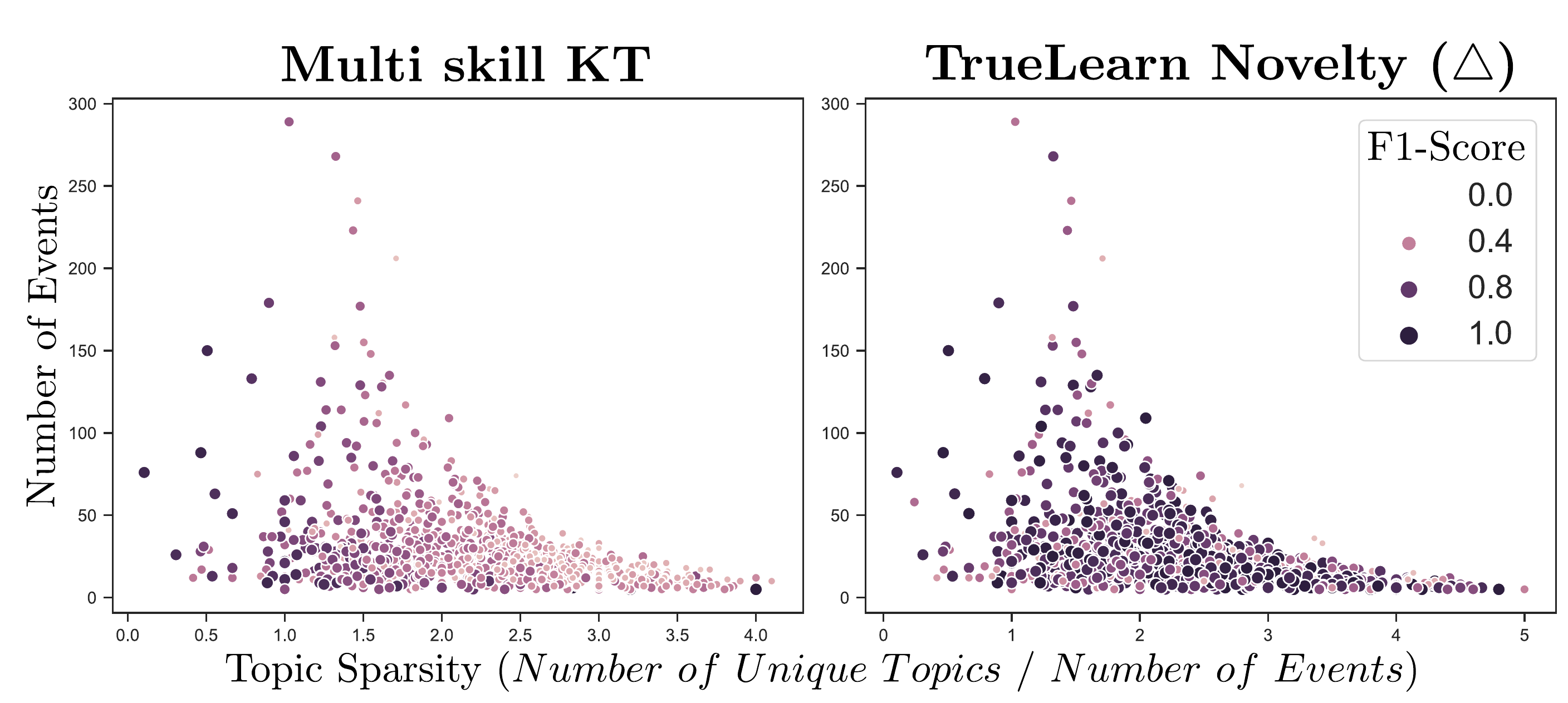}
 \caption{F1 score of each learner with their associated topic sparsity (x-axis) and number of events (y-axis). Each data point represents a learner. Colours represent F1-Score.}
 \label{tab:sparsity}
\end{figure}

We compare the approaches presented in the methodology to two {na\"ive models (persistence and majority)}. Persistence assumes that the current state of engagement will prevail, whereas majority uses the majority user engagement to decide on future engagement. We use the dataset with the 20 most active learners to validate as well the number of top $k$ topics, running the same models both for 5 and 10 topics. 
Table \ref{tab:truelearn_performance} shows the results, where highest performance for each dataset is highlighted in \textbf{bold} face and the second best in \textit{italic}. 
Firstly, we can see that the na\"ive persistence model is very competitive. This is mainly because we are predicting fragments, and persistence has an advantage in this case, as it is usually more probable that if you are engaged, you will stay engaged. However, note that the persistence will perform trivially when recommending new resources. The majority model is very competitive in terms of accuracy, as was expected. However, due to the imbalanced design of the problem we consider F1 to be a more suitable metric.  The algorithms labelled with $\bigtriangleup$ use both positive and negative engagement labels. We run these to validate our hypothesis that no assumption can be made about negative engagement unless using an engagement margin (as shown in \figurename{ \ref{fig:hypotheses}}). Both types of models achieve very similar performance in the case of Multi skill KT. In the case of TrueLearn fixed-depth it is better not to use negative engagement. This goes in line with our assumption. We also validate cosine similarity as a proxy for knowledge depth, as  TrueLearn fixed-depth achieves better performance than TrueLearn dynamic-depth, which is run for the whole dataset and infers the latent knowledge depth. The results also show very similar or improved performance when using $5$ topics, which is why we use it for the dataset containing all learners (18,933 users). For F1, TrueLearn-based models beat the baselines in most cases and achieve very promising performance, with TrueLearn Novelty achieving the most competitive performance. We thus validate the necessity of considering novelty, matching the knowledge state of learners and resources. Note that in this case, TrueLearn can make use of negative engagement, given our assumption in \figurename{ \ref{fig:hypotheses}}. Finally, \figurename{ \ref{tab:sparsity}} compares how F1 score changes for individual learners with respect to number of events and topic sparsity. It is evident that KT model struggles with learners that have high topic sparsity (with F1 score of 0 for users with topic sparsity $>$ 4). However, this is not the case for TrueLearn (similar results are obtained for other TrueLearn versions).

\paragraph{Vanilla TrueSkill Video vs TrueLearn} TrueLearn Novelty is seen to achieve similar performance to Vanilla TrueSkill Video, which considers only one skill for each learner and resource. This is as opposed to Vanilla TrueSkill, which models one skill per learner and resource fragment and achieves significantly worse performance than both. More specifically, TrueLearn Novelty is better for the 20 most active users and Vanilla TrueSkill Video for the complete dataset (which includes users with very few events). We believe this might be because: i) Vanilla TrueSkill Video can be thought as addressing how engaging resources are by only modelling one skill for the whole resource, rather than learner's knowledge and ii) the sparsity of the representation in TrueLearn might play a role in its low performance for users with few sessions. By analysing the results closely, we validated that Vanilla TrueSkill Video has better performance at the beginning of the user session, when very few user data has been collected. However, when more data is given, TrueLearn achieves better performance. To validate this, we compared the mean F1 performance for the test set of the 20 most active users at two different stages: i) the average performance over the first 100 events of a user session and ii) the average performance over the last 100 events of a user session. For i), Vanilla TrueSkill Video achieved 0.855 whereas TrueLearn Novelty 0.853, Vanilla TrueSkill Video showing slightly better performance at the beginning of the user session. For ii), Vanilla TrueSkill Video achieved 0.783 whereas TrueLearn Novelty 0.796, i.e. TrueLearn obtained similar or better performance while providing interpretable information about the knowledge state of the learner and being a more scalable solution that is run per learner (as opposed to a solution that needs to be run for the whole population and set of resources).

\paragraph{Discussion on desired features}
Regarding the desired features outlined in the introduction, we have proposed i) a transparent learner representation, that represents knowledge as a human interpretable multi-dimensional Gaussian random variable in Wikipedia topic space, key to promote self-reflection in learners \cite{Bull2016}; ii) a scalable online model that is run per user; iii) the use of an automatic content analytics toolbox that allows for cross-modality and cross-linguality features and iv) a learning strategy to make use of implicit signals in the form of learner's engagement for video lectures. In summary, TrueLearn Novelty presents superior results to the rest of models, beating Vanilla TrueSkill Video in the long run while maintaining a richer learner representation and being more scalable.

\section{Conclusions}

This work sets the foundations towards building a lifelong learning recommendation system for education. We present three different approaches, inspired by Item Response Theory and Knowledge Tracing. Our proposed model (TrueLearn) introduces the concept of novelty as a function of learner engagement. 
In this framework, recommendation algorithms need to focus on making recommendations for which i) the learner has enough background knowledge so they are able to understand and learn from the recommended material, and ii) the material has enough novelty that would help the learner to improve their knowledge about the subject. 
Our results using a very large dataset show the potential of such an approach. TrueLearn also embeds scalability, transparency and data efficiency in the core of its design showing clear promise towards building an effective lifelong learning recommendation system.
While there has been vast amount of work in context of recommendation, recommendation in education has unique challenges, due to which most existing recommendation algorithms tend not to be directly applicable.
Because of this, the list of future work remains extensive. Concerning the model: 
i) We believe that the use of a hierarchical approach, that takes into account Wikipedia link graph and category tree (i.e. dependency and correlations between KCs), could significantly improve the results by alleviating the sparsity of the KCs selected by Wikifier. This might also allow for more refined definitions of novelty. ii) The model could also be extended to consider explicit learner's feedback of the type "too difficult" or "too easy". iii) The model could be combined with a form of collaborative filtering to tackle learners with limited number of sessions. 
Concerning the data and the validation: 
i) We would like to validate our strategy for processing resources in a cross-modal and cross-lingual environment, to further strengthen our experimental analysis.
ii) We plan to set a user study to validate the recommendations of the system and design a visualisation of the learner's knowledge.

\section{Acknowledgments}
 The research described in this paper was conducted as part of the X5GON project (www.x5gon.org) which has received funding from the European Union's Horizon 2020 research and innovation programme under grant agreement No 761758.

\bibliography{aaai}
\bibliographystyle{aaai}
\end{document}